\documentclass[10pt,twocolumn,letterpaper]{article}

\usepackage{iccv}              

\usepackage{pifont}
\definecolor{iccvblue}{rgb}{0.21,0.49,0.74}
\usepackage[pagebackref,breaklinks,colorlinks,allcolors=iccvblue]{hyperref}
\usepackage[accsupp]{axessibility}

\def\paperID{2056} 
\def\confName{ICCV}
\def\confYear{2025}

\title{RapVerse: Coherent Vocals and Whole-Body Motion Generation from Text}

\author{
Jiaben Chen$^{1}$ \quad
Xin Yan$^{2}$ \quad
Yihang Chen$^{3}$ \quad
Siyuan Cen$^{1}$ \quad
Zixin Wang$^{1}$ \quad 
Qinwei Ma$^{4}$ \quad \\
Haoyu Zhen$^{1}$ \quad
Kaizhi Qian$^{5}$ \quad
Lie Lu$^{6}$ \quad
Chuang Gan$^{1,5}$ \quad
\\
\vspace{-0.5em}
\\
\small$^1$UMass Amherst\quad
$^2$Wuhan University\quad
$^3$UC San Diego\quad
$^4$Tsinghua University\quad\\
\small$^5$MIT-IBM Watson AI Lab\quad
$^6$Dolby Laboratories\quad
}

\begin{document}
\twocolumn[{
\renewcommand\twocolumn[1][]{#1}
\maketitle
\begin{center}
    \includegraphics[width=0.8\linewidth]{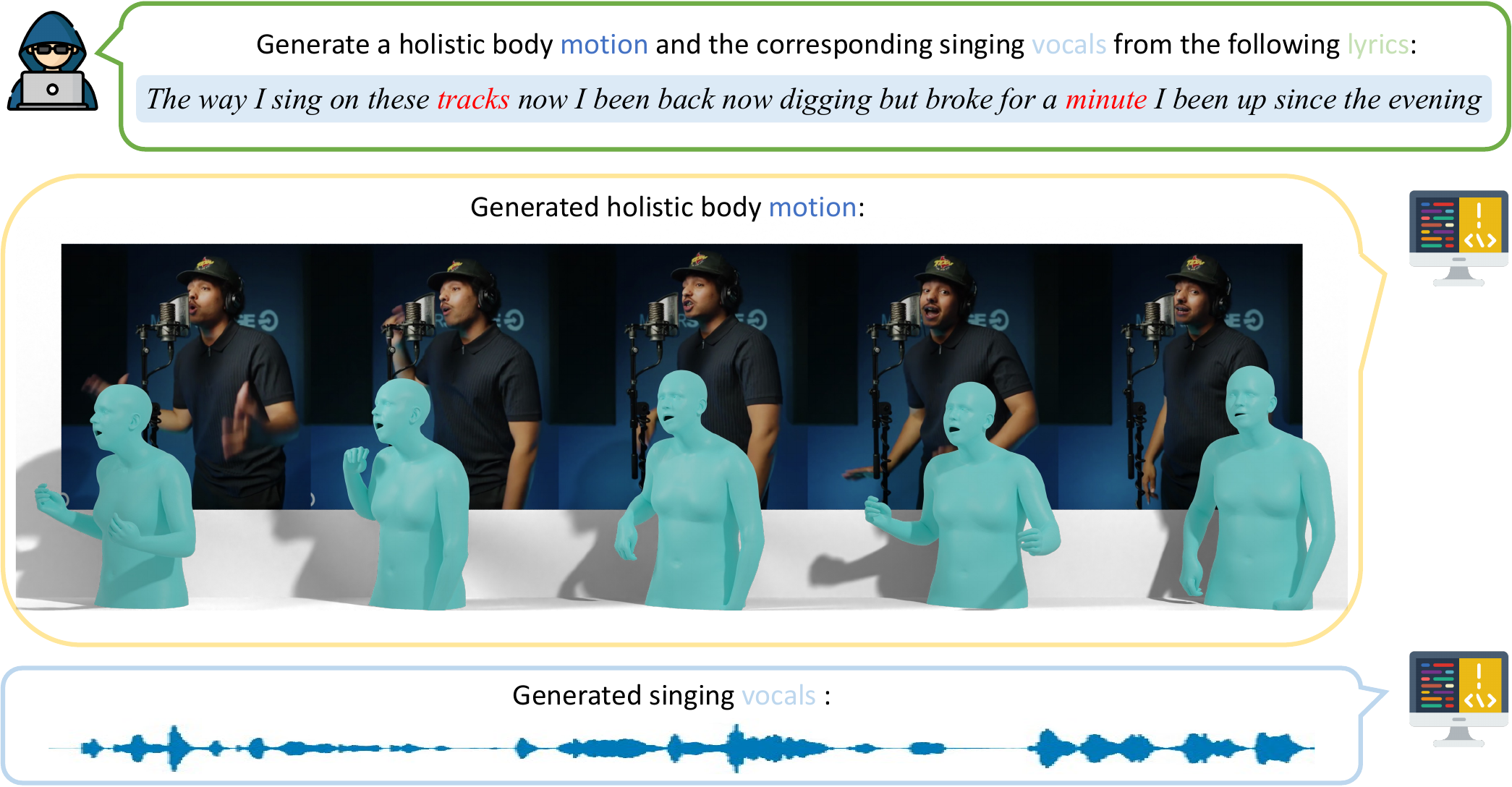}
    \captionof{figure}{\textbf{RapVerse.} We present a unified text-conditioned multi-modality generation framework, for jointly generating holistic body motions and singing vocals from textual lyrics inputs only. Note that the corresponding video frames are just shown for reference.}\label{fig:teaser}
\end{center}
}]


\begin{abstract}
In this work, we introduce a challenging task for simultaneously generating 3D holistic body motions and singing vocals directly from textual lyrics inputs, advancing beyond existing works that typically address these two modalities in isolation. To facilitate this, we first collect the RapVerse dataset, a large dataset containing synchronous rapping vocals, lyrics, and high-quality 3D holistic body meshes. With the RapVerse dataset, we investigate the extent to which scaling autoregressive multimodal transformers across language, audio, and motion can enhance the coherent and realistic generation of vocals and whole-body human motions. For modality unification, a vector-quantized variational autoencoder is employed to encode whole-body motion sequences into discrete motion tokens, while a vocal-to-unit model is leveraged to obtain quantized audio tokens preserving content, prosodic information and singer identity. By jointly performing transformer modeling on these three modalities in a unified way, our framework ensures a seamless and realistic blend of vocals and human motions. Extensive experiments demonstrate that our unified generation framework not only produces coherent and realistic singing vocals alongside human motions directly from textual inputs, but also rivals the performance of specialized single-modality generation systems, establishing new benchmarks for joint vocal-motion generation. Video demonstration and more information can be found on
the project page\footnotemark.
\end{abstract}
\footnotetext{\url{https://jiabenchen.github.io/RapVerse/}.}
\section{Introduction \label{sec:intro}}

In the evolving landscape of multi-modal content generation in terms of sound and motion, significant strides have been made in individual modalities, including text-to-music \cite{schneider2023mo, agostinelli2023musiclm, huang2023noise2music}, text-to-vocal \cite{liu2022diffsinger}, text-to-motion \cite{guo2022generating, zhang2023t2m, chen2023executing, jiang2024motiongpt, lu2023humantomato}, and audio-to-motion
\cite{yi2023generating,habibie2021learning,liu2023emage} generation. These developments have paved the way for creating more dynamic and interactive digital content. Despite these advancements, existing works predominantly operate in silos, addressing each modality in isolation. 
However, there's strong psychological evidence that for human beings, the generation of sound and motion are highly related and coupled \cite{lakoff1980metaphors}. A unified system for joint generation allows for a more expressive and nuanced communication of emotions, intentions, and context, where the generation of one modality could guide and assist the other in a coherent and efficient way.

In this paper, we tackle a crucial problem: can a machine not only sing with emotional depth but also perform with human-like expressions and motions?
We propose a novel task for generating coherent singing vocals and whole-body human motions (including body motions, hand gestures, and facial expressions) simultaneously, see Fig. \ref{fig:teaser}. This endeavor holds practical significance in fostering more immersive and naturalistic digital interactions, thereby elevating virtual performances, interactive gaming, and the realism of virtual avatars. 

An important question naturally arises: what constitutes a good model for unified generation of sound and motion? 
Firstly, we consider textual lyrics as the proper form of inputs for the unified system, since text provides a highly expressive, interpretable flexible means of conveying information by human beings, and could serve as a bridge between various modalities.
Previous efforts explore scores \cite{liu2022diffsinger}, action commands \cite{zhang2023t2m, chen2023executing, jiang2024motiongpt}, or audio signals \cite{yi2023generating} as inputs, which are inferior to textual inputs in terms of semantic richness, expressiveness and flexible integration of different modalities.

Secondly, we reckon that a joint generation system that could produce multi-modal outputs simultaneously is better than a cascaded system that executes the single-modal generation sequentially. 
A cascaded system, combining a text-to-vocal module with a vocal-to-motion module,   risks accumulating errors across each stage of generation. For instance, a misinterpretation in the text-to-vocal phase can lead to inaccurate motion generation, thereby diluting the intended coherence of the output. Furthermore, cascaded architectures necessitate multiple training and inference phases across different models, substantially increasing computational demands.

To build such a joint generation system, the primary challenges include: 1) the scarcity of datasets that provide lyrics, vocals, and 3D whole-body motion annotations simultaneously; and 2) the need for a unified architecture capable of coherently synthesizing vocals and motions from text. In response to these challenges, we have curated \textbf{RapVerse}, a large-scale dataset featuring a comprehensive collection of lyrics, singing vocals, and 3D whole-body motions. Despite the existence of datasets available for text-to-vocal \cite{liu2022diffsinger, huang2021multi, duan2013nus, sharma2021nhss}, text-to-motion \cite{plappert2016kit, mahmood2019amass, guo2022generating, lin2024motion}, and audio-to-motion \cite{cao2017realtime, habibie2021learning, ginosar2019learning, fanelli20103, cudeiro2019capture, wuu2022multiface}, the landscape lacks a unified dataset that encapsulates singing vocals, whole-body motion, and lyrics simultaneously. Most notably, large text-to-vocal datasets \cite{huang2021multi, zhang2022m4singer} are predominantly in Chinese, limiting their applicability for English language research and lacking any motion data. And text-to-motion datasets \cite{plappert2016kit, guo2022generating, lin2024motion} typically focus on text descriptions of specific actions paired with corresponding motions without audio data, often not covering whole body movements. Moreover, audio-to-motion datasets \cite{liu2022diffsinger, lu2023co} focus primarily on speech rather than singing. A comparison of existing related datasets is demonstrated in Table. \ref{tab:datasets_comparison}. The RapVerse dataset is divided into two distinctive parts to cater to a broad range of research needs: 1) a Rap-Vocal subset containing a large number of pairs of vocals and lyrics, and 2) a Rap-Motion subset encompassing vocals, lyrics, and human motions. The Rap-Vocal subset contains 108.44 hours of high-quality English singing voice in the rap genre without background music. Paired lyrics and vocals are crawled from the Internet from 32 singers, with careful cleaning and post-processing. On the other hand, the Rap-Motion subset contains 26.8 hours of rap performance videos with 3D holistic body mesh annotations in SMPL-X parameters \cite{pavlakos2019expressive} using the annotation pipeline of Motion-X \cite{lin2024motion}, synchronous singing vocals and corresponding lyrics. 

With the RapVerse dataset, we explore how far we can push by simply scaling autoregressive multimodal transformers with language, audio, and motion for a coherent and realistic generation of vocals and whole-body human motions. To this end, we unify different modalities as token representations. Specifically, three VQVAEs \cite{van2017neural} are utilized to compress whole-body motion sequences into three-level discrete tokens for head, body, and hand, respectively. For vocal generation, previous works \cite{min2021meta, donahue2020end, liu2022diffsinger, min2021meta} share a common paradigm, producing mel-spectrograms of audio signals from input textual features and additional music score information, following with a vocoder \cite{oord2016wavenet, valin2019lpcnet, yang2021multi} to reconstruct the phase. We draw inspiration from the speech resynthesis domain \cite{polyak2021speech}, and learn a self-supervised discrete representation to quantize raw audio signal into discrete tokens while preserving the vocal content and prosodic information. Then, with all the inputs in discrete representations, we leverage a transformer to predict the discrete codes of audio and motion in an autoregressive fashion. 
Extensive experiments demonstrate that this straightforward unified generation framework not only produces realistic singing vocals alongside human motions directly from textual inputs but also rivals the performance of specialized single-modality generation systems.

To sum up, this paper makes the following contributions:
\begin{itemize}
\item We collect RapVerse, a large dataset featuring synchronous singing vocals, lyrics, and high-quality 3D holistic SMPL-X parameters.

\item We design a simple but effective unified framework for the joint generation of singing vocals and human motions from text with a multi-modal transformer in an autoregressive fashion.

\item To unify representations of different modalities, we employ a vocal-to-unit model to obtain quantized audio tokens and utilize compositional VQVAEs to get discrete motion tokens.

\item Experimental results show that our framework rivals the performance of specialized single-modality generation systems, setting new benchmarks for joint generation of vocals and motion.
\end{itemize}

\section{Related Work \label{sec:related}}
\subsection{Text to Vocal Generation}
\noindent\textbf{Text-to-audio Dataset.} There exists several singing vocal datasets, yet they each has constraints. For instance, PopCS \cite{liu2022diffsinger} and OpenSinger \cite{huang2021multi} are limited to Chinese, while NUS-48E \cite{duan2013nus} and NHSS \cite{sharma2021nhss} have only a few hours of songs. Nevertheless, JVS-MuSiC \cite{tamaru2020jvs} and NUS-48E \cite{duan2013nus} offer a few hours of songs from dozens of singers, whereas OpenSinger \cite{huang2021multi} provides a more extensive collection with tens of hours from a single singer. Notably, our dataset represents the first to be specifically curated for rap songs of multiple singers with 108 hours.

\noindent\textbf{Text-to-vocal Models.} Recent advancements in text-to-speech (TTS) models, including WavNet \cite{oord2016wavenet}, FastSpeech 1 and 2 \cite{ren2019fastspeech, ren2020fastspeech}, and EATS \cite{donahue2020end}, have significantly improved the quality of synthesized speech. However, singing voice synthesis (SVS) presents a greater challenge due to its reliance on additional musical scores and lyrics. Recent generation models \cite{liu2022diffsinger, he2023rmssinger, he2023rmssinger, zhang2022wesinger,kim2023muse} perform excellently in generating singing voices. However, their performance tends to deteriorate when encountering out-of-distribution data. To handle this, StyleSinger \cite{zhang2023stylesinger} introduces a Residual Style Adaptor and an Uncertainty Modeling Layer Normalization to handle this. 

\newcommand{\cmark}{\ding{51}}%
\newcommand{\xmark}{\ding{55}}%
\begin{table*}[t]
\scriptsize
\begin{center}    
\subfloat[
Audio Datasets
]{
\hspace{-0em}
\begin{minipage}[t]{0.48\linewidth}
    \begin{center}
\begin{tabular}{l|ccc}
\toprule
Dataset & Language & \#Singers & Hours \\
\midrule
NUS-48E \cite{duan2013nus} & English & 12 & 1.91\\
NHSS \cite{sharma2021nhss} & English & 10 & 7 \\
JVS-MuSiC \cite{tamaru2020jvs} & Japanese & 100 & 2.28\\
Tohoku Kiritan \cite{ogawa2021tohoku} & Japanese & 1 & 1\\
PopCS \cite{liu2022diffsinger} & Chinese & 1 & 5.89 \\
OpenSinger \cite{huang2021multi} & Chinese & 66 & 50 \\
Opencpop \cite{wang2022opencpop} & Chinese & 1 & 5.25 \\
M4Singer \cite{zhang2022m4singer} & Chinese & 20 & 29.77\\
\midrule
\textbf{Ours} & English & 32 & 108.44 \\
\bottomrule
\end{tabular}
\end{center}
\end{minipage}
}
\hspace{0em}
\subfloat[
Motion Datasets
]{
\begin{minipage}[t]{0.48\linewidth}
    \begin{center}
\begin{tabular}{l|ccc}
\toprule
Dataset & Modality & Whole$^\dag$ & Hours \\
\midrule
KIT-ML \cite{plappert2016kit} & Text & \scalebox{0.85}[1]{\xmark} & 11.2 \\
AMASS \cite{mahmood2019amass} & Text & \scalebox{0.85}[1]{\xmark} & 40.0 \\
HumanML3D \cite{guo2022generating} & Text & \scalebox{0.85}[1]{\xmark} & 28.6 \\
Motion-X \cite{lin2024motion} & Text & \scalebox{0.85}[1]{\cmark} & 127.1 \\
HA2G \cite{liu2022diffsinger} & Audio & \scalebox{0.85}[1]{\xmark} & 33 \\
Yoon et.al \cite{lu2023co} & Audio & \scalebox{0.85}[1]{\xmark} & 30 \\
Speech2Gesture \cite{ginosar2019learning} & Audio & \scalebox{0.85}[1]{\xmark} & 144 \\
Talkshow \cite{yi2023generating} & Audio & \scalebox{0.85}[1]{\cmark} & 27.0 \\
\midrule
\textbf{Ours} & Text-Audio & \scalebox{0.85}[1]{\cmark} & 26.8 \\
\bottomrule
\end{tabular}
    \end{center}
\end{minipage}
}
\end{center}
\vspace{-15pt}
\caption{\textbf{Comparison of Audio and Motion Datasets.} (a) compares our Rap-Vocal Subset with existing singing vocal datasets. (b) compares our Rap-Motion Subset with existing motion datasets. $^\dag$: ``Whole'' means whole-body motion including face, body, and hand. }
\vspace{-5mm}
\label{tab:datasets_comparison}
\end{table*}

\begin{figure*}[htbp]
    \centering
    \includegraphics[width=0.5\linewidth]{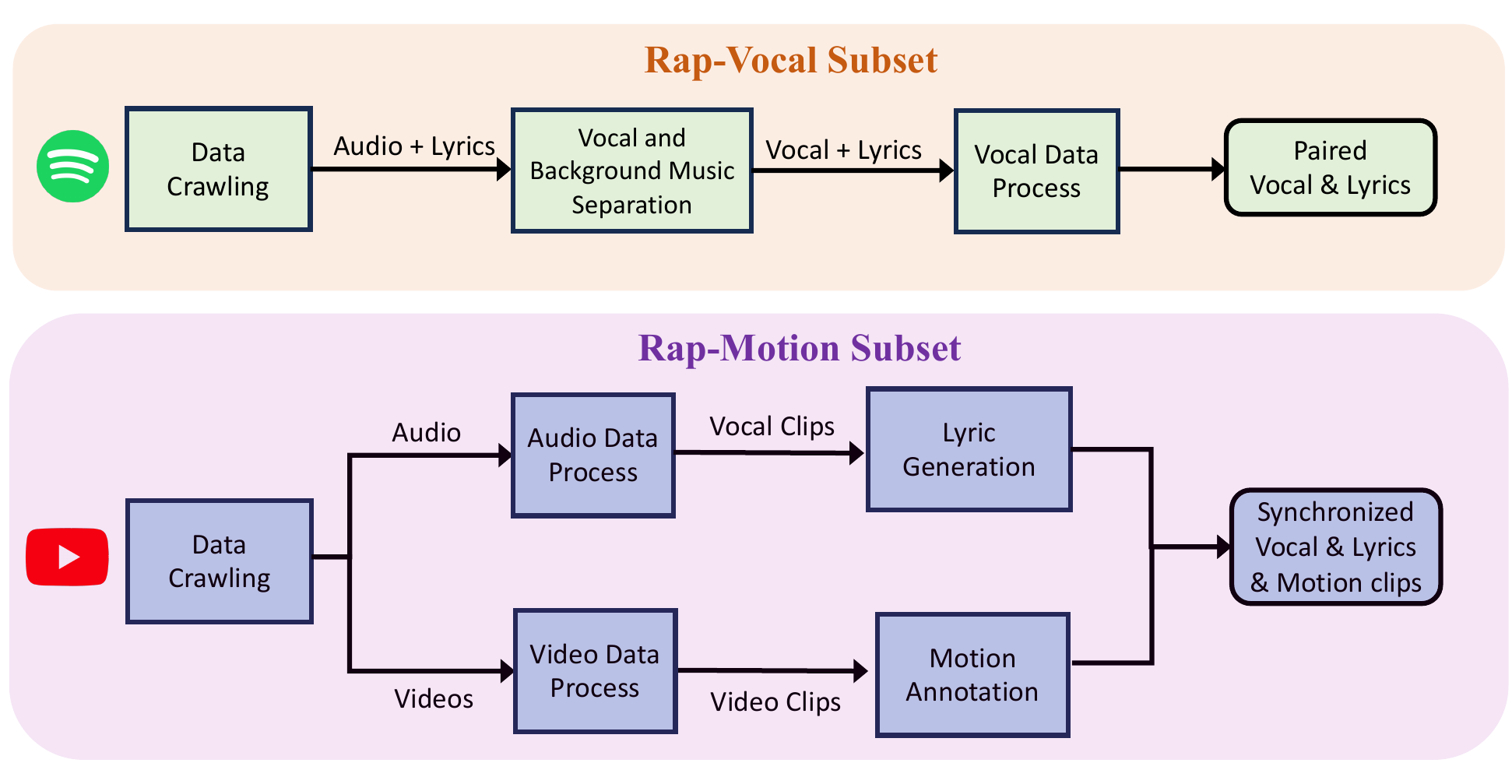}
    \vspace{-5pt}
    \caption{\textbf{RapVerse} dataset collection pipeline. There are two pathways for rap-vocal and rap-motion subsets, respectively.}
    \label{fig:audio-data}
\end{figure*}

\subsection{Text to Motion Generation}
\noindent\textbf{Text-to-motion Dataset.} Current text-motion datasets, such as KIT \cite{plappert2016kit}, AMASS \cite{mahmood2019amass}, and HumanML3D \cite{guo2022generating}, are constrained by limited data coverage, typically spanning only tens of hours and lacking whole-body motion representation. To overcome these shortcomings, Motion-X \cite{lin2024motion} was developed, providing an extensive dataset that encompasses whole-body motions.\\ 
\noindent\textbf{Text-to-motion Models.} Text-to-motion models \cite{petrovich2022temos, zhang2024motiondiffuse, guo2022tm2t, guo2022generating, lu2023humantomato, chen2023executing} have gained popularity for their user convenience. Recent advancements have focused on diffusion models \cite{zhang2024motiondiffuse, kim2023flame, tevet2022human}, which, unlike deterministic generation models \cite{ahuja2019language2pose, ghosh2021synthesis}, enable finer-grained and diverse output generation. Chen et al. \cite{chen2023executing} proposed a motion latent-based diffusion model to enhance generative quality and reduce computational costs. However, the misalignment between natural language and human motions presents challenges. Lu et al. \cite{lu2023humantomato} introduced the Text-aligned Whole-body Motion generation framework to address these issues by generating high-quality, diverse, and coherent facial expressions, hand gestures, and body motions simultaneously.

\subsection{Audio to Motion Generation}
\noindent\textbf{Audio-to-motion Dataset.} Co-Speech Datasets can be classified into 1: pseudo-labeled (PGT) and 2: motion-captured (mocap). PGT datasets \cite{cao2017realtime, habibie2021learning, ginosar2019learning, yi2023generating} utilize disconnected 2D or 3D key points to represent the body, allowing extracting at a lower cost but with limited accuracy. On the other hand, mocap datasets provide annotations for limited body parts, with some focusing solely on the head \cite{fanelli20103, cudeiro2019capture, wuu2022multiface} or the body \cite{takeuchi2017creating, ferstl2018investigating}. Differing from them, BEATX \cite{liu2023emage} contains mesh data of both head and body. However, these datasets typically focus on speech to motion generation. Our RapVerse, in contrast, contains paired text-vocal-motion data, enabling simultaneous motion and vocal generation.

\noindent\textbf{Audio-to-motion Models.}  Audio-to-motion models \cite{taylor2017deep, shen2023difftalk, zhang2023sadtalker, liu2023emage, yi2023generating, alexanderson2023listen, bian2024adding, chen2024diffsheg} aim to produce human motion from audio innputs. Recognizing the intricate relationship between audio and human face, TalkSHOW \cite{yi2023generating} separately generates face parts and other body parts. However, this approach exhibits limitations, such as the absence of generating lower body motion. Motioncraft \cite{bian2024adding} proposes a DiT structure for motion generation. EMAGE \cite{liu2023emage} utilizes masked gestures together with vocals for a generation. Our model, however, goes a step further by being the first to generate paired audio-motion data directly from text. 
\section{RapVerse Dataset \label{sec:dataset}}
In this section, we introduce RapVerse, a large rap music motion dataset containing synchronized singing vocals, textual lyrics and whole-body human motions. 
A comparison of our dataset with existing datasets is shown at Table. \ref{tab:datasets_comparison}. The RapVerse dataset is divided into two subsets to cater to a broad range of research needs: a Rap-Vocal subset and a Rap-Motion subset. The overall collection pipeline of RapVerse is shown at Fig. \ref{fig:audio-data}.
\vspace{-4pt}
\subsection{Rap-Vocal Subset}
The Rap-Vocal subset contains 108.44 hours of high-quality English singing voice in the rap genre with paired lyrics. We will introduce each step in detail.

\noindent\textbf{Data Crawling.} In a bid to obtain a large number of rap songs and corresponding lyrics from the Internet, we utilize Spotdl and Spotipy to collect songs, lyrics, and metadata of different rap singers. To ensure the quality of the dataset, we perform cleaning on the crawled songs by removing songs with misaligned lyrics and filtering out songs that are too long or too short.

\noindent\textbf{Vocal and Background Music Separation.} Since the crawled songs are mixed with rapping vocals and background music, and we aim to synthesize singing vocals from separated clean data, we utilize Spleeter \cite{hennequin2020spleeter}, the state-of-the-art open-source vocal-music source separation tool to separate and extract rap vocal voices and accompanying background music from the collected songs. Following \cite{ren2020deepsinger}, we normalize the loudness of the vocal voices to a fixed loudness level.

\noindent\textbf{Vocal Data Processing.} The raw crawled lyrics from the Internet are in inconsistent formats, we conduct data cleaning on the lyrics by removing meta information (singer, composer, song name, bridging words, and special symbols). To ensure that the lyrics are aligned with the singing vocals, we collect lyrics only with the correct timestamps of each sentence, and we separate each song into around 10-second to 20-second segments for model training. 

\subsection{Rap-Motion Subset}
The Rap-Motion subset contains 26.8 hours of rap performance videos with 3D holistic body mesh annotations in SMPL-X parameters \cite{pavlakos2019expressive}, synchronous singing vocals, and corresponding lyrics. We introduce the collection pipeline of this subset as follows.

\noindent\textbf{Data Crawling.} We crawled over 1000 studio performance videos from YouTube under the Common Creative License. We filter out low-quality videos manually to ensure the videos meet the following criteria: stable camera work, performers centered in the frame, clear visibility of the performer's entire body to capture detailed motion data, and high-quality audio for accurate vocal analysis. 

\noindent\textbf{Audio Data Processing.} Similar to the Rap-Vocal subset, we leverage Spleeter \cite{spleeter2020} to isolate singing vocals from accompanying music. Given that YouTube videos typically lack paired lyrics, we employ an ASR model, Whisper \cite{radford2023robust}, to accurately transcribe vocals into corresponding text.

\noindent\textbf{Video Data Processing.} To ensure the collection of high-quality video clips for motion annotation, we implemented a semi-automatic process to filter out undesirable content, such as advertisements, transition frames, changes in shots, and flashing lights. Initially, we applied YOLO \cite{redmon2018yolov3} for human detection to discard frames where no humans were detected. Subsequently, we utilized RAFT \cite{teed2020raft} to assess the motion magnitude, employing a threshold to eliminate frames affected by camera instability. We then perform meticulous manual curation on the extracted clips, retaining only those of the highest quality. Finally, we follow the pipeline of the optimized-based method Motion-X \cite{lin2024motion} to extract 3D whole-body meshes from monocular videos. Specifically, we adopt the SMPL-X \cite{pavlakos2019expressive} for motion representations,  given a T-frame video clip, the corresponding pose states $\mathcal{M}$ are represented as:
\begin{equation}
    \mathcal{M} = \{\mathcal{M}_f, \mathcal{M}_b, \mathcal{M}_h, \zeta, \epsilon \}
\end{equation}
where $\mathcal{M}_f \in \mathbb{R}^{T\times3}, \mathcal{M}_b \in \mathbb{R}^{T\times63}$ and $\mathcal{M}_h \in \mathbb{R}^{T\times90}$ denotes jaw poses, body poses and hand poses, respectively. $\zeta \in \mathbb{R}^{T\times100}$ and $\epsilon \in \mathbb{R}^{T\times3}$ are the facial expression and global translation.
\begin{figure*}[t]
    \centering
    \includegraphics[width=0.7\linewidth]{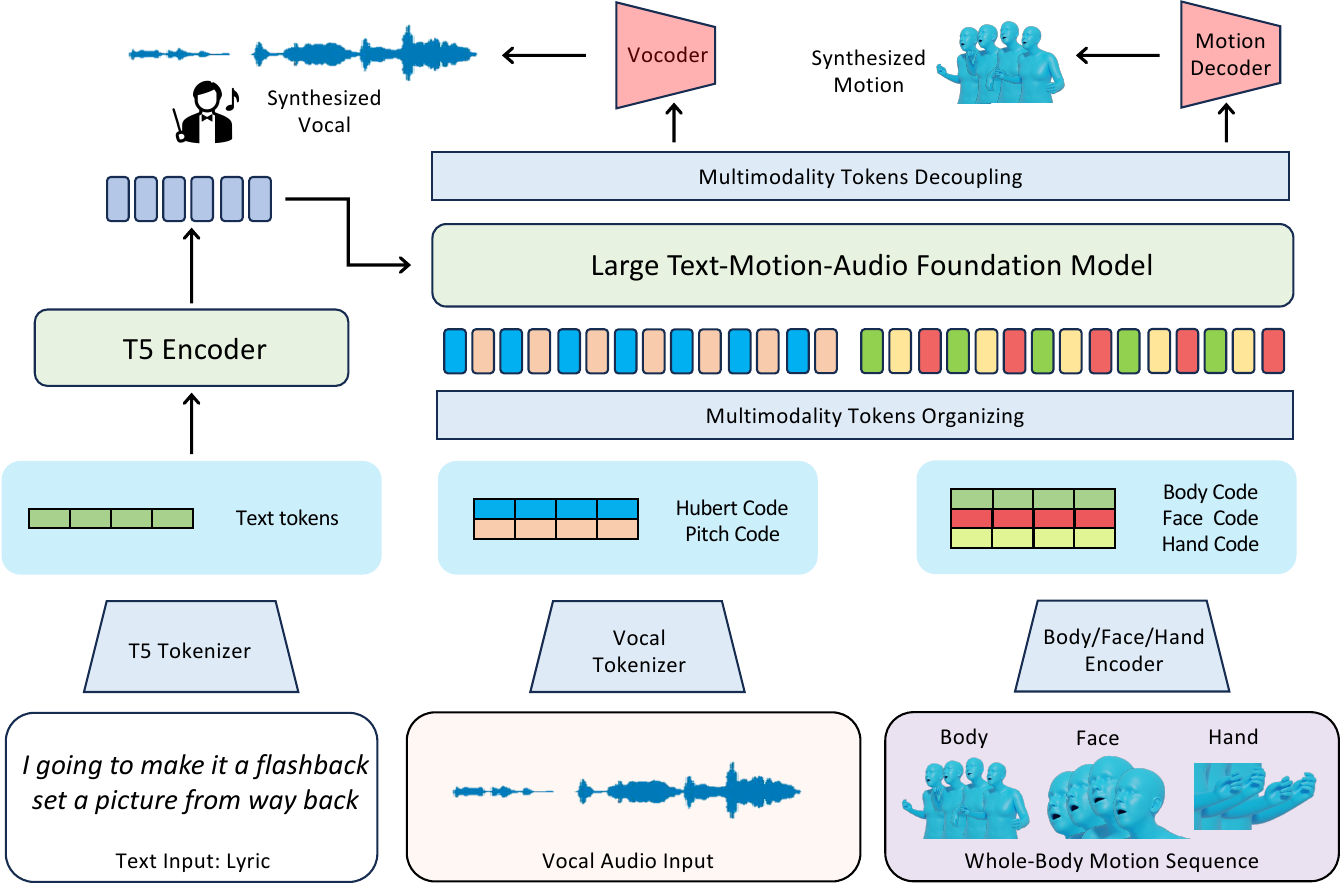}
    \caption{\textbf{Pipeline overview.} We first pre-train all tokenizers on vocal-only and motion-only data. After we have pretrained the modality tokenizers, we can unify text, vocal, and motion in the same token space. We adopt a mixing organizing algorithm for input tokens to align via the temporal axis. These mixed input tokens are fed into the large Text-Motion-Audio foundation model to train on token prediction tasks, guided by the encoded features from textual input.}
    \label{fig:pipeline}
\end{figure*}

\section{Method \label{sec:method}}

Given a piece of lyrics text, our goal is to generate rap-style vocals and whole-body motions, including body movements, hand gestures, and facial expressions that resonate with the lyrics. With the help of our RapVerse dataset, we propose a novel framework that not only represents texts, vocals, and motions as unified token forms but also integrates token modeling in a unified model. As illustrated in Fig.~\ref{fig:pipeline}, our model consists of multiple tokenizers for motion (Sec.~\ref{sec:motion_tokenizer}) and vocal (Sec.~\ref{sec:audio_tokenizer}) token conversions, as well as a general Large Text-Motion-Audio Foundation Model (Sec.~\ref{sec:transformer}) that targets for audio token synthesize and motion token creation, based on rap lyrics.

\subsection{Problem Formulation}\label{sec:problem}

Let $\mathcal{L}$ be the rap lyric, such as ``\textit{I going to make it a flashback set a picture from way back.}'', our model will compose the text-related vocal $\mathcal{V}\in \mathbb{R}^{t\times d_v}$ and whole-body motion $\mathcal{M}\in \mathbb{R}^{t\times d_m}$, where $t$, $d_v$, $d_m$ denote the temporal length and the feature dimensions for vocal and motion in every time unit, respectively. Modality tokenizer encoders $\phi^\mathcal{X}(\mathcal{T}\mid \mathcal{X})$ and decoders $\varphi^\mathcal{X}(\mathcal{X}\mid \mathcal{T})$ will bridge each modality with the token format $\mathcal{T}$ utilizing different tokenizers, where $\mathcal{X} \in \{\mathcal{L}, \mathcal{V}, \mathcal{M}\}$. The large foundation model $\psi^\mathcal{X}$ will incorporate tokens in unified modeling. The whole problem of coherent vocal and model generation via lyric text can be formulated as follows:
\begin{equation}
\Theta^* = \mathop{\arg\max}_{\Theta} \mathbb{P}_{\Theta}(\mathcal{V}, \mathcal{M} \mid \mathcal{L}),
\end{equation}
where $\Theta$ denotes the whole model parameters and $\mathbb{P}_{\Theta}(\cdot)$ denotes the model distributions.

\subsection{Motion VQ-VAE Tokenizer}\label{sec:motion_tokenizer}

Similar to prior arts, a motion VQ-VAE is adopted to convert 3D human motions into discrete tokens, where the tokenizer encoder $\phi$ generates motion tokens enriched with high-frequency information, while the decoder $\varphi$ seamlessly reconstructs discrete tokens into continuous motion sequences. Specifically, the tokenizer will maintain a learnable codebook $\mathcal{C}$, which firstly encodes the input motion features $\mathcal{M}$ to vectors $z^\mathcal{M}$, and then the motion quantizer $\mathcal{Q}^\mathcal{M}$ will look up the nearest neighbor in codebook $\mathcal{C}^\mathcal{M}=\{c_k^\mathcal{M}\}_{k=1}^{d_{c}}$ for the motion representation. The quantized code can be calculated as
\begin{equation}
Q(z^\mathcal{M};\mathcal{C}^\mathcal{M}) = \mathop{\arg\min}_{k} \|z^\mathcal{M} - c_k^\mathcal{M}\|_2.
\end{equation}
The tokenizer decoder $\varphi$ will project the quantized code back into the human motion after quantization. The reconstruction objective in the tokenizer training stage can be represented as
\begin{equation}
\mathbb{E}_{\mathcal{T}\sim\phi_\mathcal{M}(\mathcal{T}\mid \mathcal{M})}\left[\log \varphi_\mathcal{M}(\mathcal{M}\mid \mathcal{T})\right].
\end{equation}

 As the whole-body motion $\mathcal{M}$ can be decomposed into face $\mathcal{M}_f$, body $\mathcal{M}_b$, and hand $\mathcal{M}_h$, where $\mathcal{M}\in \mathbb{R}^{t\times d}, d_f+d_b+d_h = d_m$, we build three tokenizers to obtain tokens for these three parts.

\subsection{Vocal2unit Audio Tokenizer}\label{sec:audio_tokenizer}
Overall, we leverage the self-supervised framework \cite{polyak2021speech} in speech resynthesis domain to learn vocal representations from the audio sequences. Specifically, we train a Vocal2unit audio tokenizer to build a discrete tokenized representation for the human singing voice. The vocal tokenizer consists of three encoders and a vocoder. The encoders include three different parts: (1) the semantic encoder; (2) the F0 encoder; and (3) the singer encoder. We will introduce each component of the model separately.

\noindent\textbf{Semantic Encoder.} In order to extract the semantic information of the audio into discrete tokens, we use a pre-trained Hubert encoder \cite{hsu2021hubert} as the main component of our semantic encoder network $E^{h}$. Given a raw audio sequence $v$, the output of the Hubert encoder is a continuous representation. Then we apply a K-means algorithm on the representations to finally generate discrete units $z^{\mathcal{S}} = \{z^{\mathcal{S}}_i\}_{i=1}^{L_s}$ where $z^{\mathcal{S}}_i \in \{1,2,\cdots,K\}$. $L_s$ is the length of hubert token sequence, and $K$ represents the number of clusters in K-means algorithm.

\noindent\textbf{F0 Encoder.} We use the YAAPT algorithm \cite{kasi2002yet} to extract the F0 from the input audio signal $v$, which is then fed into an encoder $E_p$ to generated low-frequency F0 pitch representation. The encoder $E^p$ is trained using a VQ-VAE framework with Exponential Moving Average updates similar to \cite{dhariwal2020jukebox}, maintaining a learnable codebook $\mathcal{C}^p=\{c_k^p\}_{k=1}^{d_{c}}$, where $d_c$ is the number of codes in the codebook. The raw F0 sequence is first preprocessed according to the singer, converted into a sequence of latent vectors, then mapped to its nearest neighbors in the codebook $c^p$, which could be represented using integer number $k\in\{1, 2,\cdots,d_c\}$ as its index in the codebook. The index sequence will be further formatted as the token \textit{<pitch\_k>} to produce pitch token sequence $z^{\mathcal{P}} = \{z^{\mathcal{P}}_i\}_{i=1}^{L_p}$, where $L_p$ is the length of the pitch token sequence. Finally, the decoder will reconstruct the F0 representation leveraging the pitch token sequence. 

\noindent\textbf{Singer Encoder.} As the last component of our encoders, the singer encoder $E_{si}$ from \cite{heigold2016end,polyak2021speech} is used to extract the Mel-spectrogram from the raw audio sequence and to output a singer representation $z^{\mathcal{I}} \in \mathbb{R}^{256}$. The singer embedding only depends on the singer and is trained globally.

\noindent\textbf{Vocoder.} In order to decode the vocal signal from the discrete tokens, we adopt a modified version of the HiFi-GAN neural vocoder similar to the one in \cite{polyak2021speech}, consisting of a generator and multiple discriminators. 
The generator takes the encoded discrete representation $z^{\mathcal{S}}$ and $z^{\mathcal{P}}$, and the singer embedding vector $z^{\mathcal{I}}$ as inputs. After a lookup table and a series of blocks composed of transposed convolution and a residual block with dilated layers, it finally outputs waveform audio. The discriminators include multiple sub-discriminators, analyzing on different scales and different periodic structures.

\subsection{General Auto-regressive Modeling}\label{sec:transformer}
\noindent\textbf{Model Architectures.}
The whole model of our general modeling pipeline consists of several pre-trained vocal and motion tokenizers and a large Text-Motion-Audio foundation model.
The vocal and motion tokenizers are based on VQ-VAE architectures, enabling us to represent these two modalities as discrete tokens. 
We also adopt a T5-Tokenizer \cite{raffel2020exploring} to convert lyric text into tokens, so that all three modalities $\mathcal{X} \in \{\mathcal{L}, \mathcal{V}, \mathcal{M}\}$ are united in the token space $\mathcal{T}^\mathcal{X}=\{t_k^\mathcal{X}\}_{k=1}^{\mathcal{N}^\mathcal{X}}$, where $\mathcal{N}^\mathcal{X}$ is the token sequence length for each modality. 
The large foundation model, a temporal transformer based on the decoder-only architecture, performs the next-token-prediction task based on textual features encoded by the T5-Encoder.

Specifically, each token is generated after the start token by the probability distribution from the foundation model $p_\delta(t^\mathcal{X};h^\mathcal{L}) = \prod\limits_ip_\delta(t_i^\mathcal{X}\mid t_{<i}^\mathcal{X};h_\mathcal{L})$. During the training period, we calculate the cross-entropy loss to measure the likelihood between ground truth tokens and predicted token probabilities as 
\begin{equation}
-\sum\limits_{i=1}^{\mathcal{N^\mathcal{X}}} \log p_\delta(t^\mathcal{X}_i;h^\mathcal{L}).
\end{equation}
After optimizing via this training objective, our model learns to predict the next token, which can be decoded into different modality features. This process is similar to text word generation in language models, while the ``word'' in our method such as \textit{<face\_0123>}, does not have explicit semantic information, but can be decoded into continuous modality features.

\noindent\textbf{Multimodality Tokens Organization.}
Since we have designed a coherent vocal and motion generation pipeline with multiple token types in our method, including \textit{hubert} and \textit{pitch} tokens in rap vocals, and \textit{face}, \textit{body}, \textit{hand} in whole-body motions, the organization method of all these tokens in one sentence is important for training. Similar to previous works~\cite{lu2023humantomato, jiang2024motiongpt}, we adopt an interleaved style within each modality. Specifically, with the input whole-body motion tokens $\mathcal{T}^{\mathcal{M}_s}=\{t^{\mathcal{M}_s}_k\}_{k=1}^{\mathcal{N}^{\mathcal{M}_s}}$, where $s=f,b,h$ which denotes face, body and hand respectively, and $\mathcal{N}^{\mathcal{M}_s}$ denotes the token length for each part. Therefore the motion token sequence can be formulated as $\mathcal{T}^{\mathcal{M}}=\{t^{\mathcal{M}_f}_1, t^{\mathcal{M}_b}_1, t^{\mathcal{M}_h}_1, t^{\mathcal{M}_f}_2,...\}$. Similarly, the vocal tokens consist of the hubert tokens $t^{\mathcal{V}_h}$ and pitch tokens $t^{\mathcal{V}_p}$, which can be formulated as $\mathcal{T}^{\mathcal{V}}=\{t^{\mathcal{V}_h}_1, t^{\mathcal{V}_p}_1,
t^{\mathcal{V}_h}_2,
t^{\mathcal{V}_p}_2,...\}$. 
This interleaved way of organization allows us to align the timesteps of each part. 
When performing coherent vocal and motion generation, we mix these two modality sequences sequentially with modality-wise start token as $\mathcal{T} = \{t_{start}^{V}, \mathcal{T}^{V}, t_{start}^{M}, \mathcal{T}^{M}\}$. Note that we put vocal sequences before motion sequences because 1) the vocals are more related to the lyrics, and 2) motion tokens have a relationship with previously generated vocals, \eg face motions with lips movement are directly related to vocals. Joint training of mixed modalities allows the transformer to capture the temporal dependencies and dynamics inherent in vocal and motion data.

\noindent\textbf{Inference and Decoupling.} In the inference stage, we use different start tokens to specify which modality to generate. The textual input is encoded as features to guide token inference. We adopt a top-k algorithm to control the diversity of the generated content by adjusting the temperature, as generating vocals and motions based on lyrics is a creation process with multiple possible answers. After token prediction, a decoupling algorithm is used to process output tokens to make sure tokens from different modalities are separated and temporally aligned. These discrete tokens will be further decoded into text-aligned vocals and motions. 
\section{Experiments \label{sec:experiment}}
In this section, we evaluate our proposed model on our proposed benchmark designed for joint vocal and whole-body motion generation from textual inputs.

\subsection{Experimental Setup}
\noindent \textbf{Metrics.} To evaluate the generation quality of singing vocals, we utilize the Mean Opinion Score (MOS) to gauge the naturalness of the synthesized vocal. For motion synthesis, we evaluate the generation quality of the body hand gestures and the realism of the face, respectively. Specifically, for gesture generation, we use Frechet Inception Distance (FID) based on a feature extractor from \cite{guo2022generating} to evaluate the distance of feature distributions between the generated and real motions, and Diversity (DIV) metric to assess the motions diversity. For face generation, we compare the vertex MSE \cite{xing2023codetalker} and the vertex L1 difference LVD \cite{yi2023generating}. Finally, we adopt Beat Constancy (BC) \cite{li2021ai} to measure the synchrony of generated motion and singing vocals. 

\noindent \textbf{Baselines.} We compare the vocal generation quality with the state-of-the-art vocal generation method DiffSinger \cite{liu2022diffsinger}. And we also adapt the text-to-speech model FastSpeech2 \cite{ren2020fastspeech} for vocal generation. For motion generation, we compare our method with both text-to-motion methods and audio-to-motion methods. For text-to-motion methods, since there is no existing open-sourced work for text to whole-body motion generation, we compare with transformer-based T2M-GPT \cite{zhang2023t2m} and MLD \cite{chen2023executing} for body generation. For the audio-to-motion generation, we compare with Habibie et al. \cite{habibie2021learning}, Talkshow \cite{yi2023generating}, Motioncraft \cite{bian2024adding} and EMAGE \cite{liu2023emage}. We report all the results on RapVerse with an 85\%/7.5\%/7.5\% train/val/test split.

\begin{table*}[t]
\small
\begin{center}    
\subfloat[
Motion Generation
]{
\begin{minipage}[t]{0.63\linewidth}
    \begin{center}
\begin{tabular}{l|cc|c|cc}
\toprule
Method & FID$\downarrow$ & DIV$\uparrow$ & BC$\uparrow$ & MSE$\downarrow$ & LVD$\downarrow$\\
\midrule
\multicolumn{6}{l}{\underline{\textit{Text-to-Motion}}} \\
T2M-GPT \cite{zhang2023t2m} & 23.45 & 11.75 & - & - & -\\
MLD \cite{chen2023executing} & 26.34 & 12.15 & - & - & -\\
\midrule
\multicolumn{6}{l}{\underline{\textit{Audio-to-Motion}}}\\
Habibie et al. \cite{habibie2021learning} & 32.14 & 10.08 & 0.476 & 2.13 & 9.54\\
Talkshow \cite{yi2023generating} & 18.23 & 13.14 & 0.482 & 2.05 & 9.20\\
EMAGE \cite{liu2023emage} & 21.18 & 12.65 & \textbf{0.488} & \textbf{1.96} & \underline{8.45}\\
MotionCraft \cite{bian2024adding} & \underline{17.75} & \underline{13.76} & 0.482 & 2.06 & 9.23\\
\midrule
Cascaded Result & 23.42 & 12.87 & 0.479 & 2.09 & 9.38\\
\midrule
\multicolumn{6}{l}{\underline{\textit{Text-to-Audio+Motion}}}\\
\textbf{Ours} & \textbf{17.58} & \textbf{14.08} & \underline{0.485} & \underline{2.03} & \textbf{7.23}\\
\bottomrule
\end{tabular}
\end{center}
\end{minipage}
}
\subfloat[
Vocal Generation
]{
\begin{minipage}[t]{0.28\linewidth}
    \begin{center}
\vspace{-38pt}
\begin{tabular}{l|c}
\toprule
Method & MOS$\uparrow$ $^\dag$\\
\midrule
GT & 4.45 ± 0.06 \\
Reconstruction & 4.02 ± 0.08 \\
\midrule
FastSpeech2 \cite{ren2020fastspeech} & 3.41 ± 0.18 \\
DiffSinger \cite{liu2022diffsinger} & 3.72 ± 0.12 \\
\midrule
\textbf{Ours} & 3.64 ± 0.15 \\
\bottomrule
\end{tabular}
    \end{center}
\end{minipage}
}
\end{center}
\vspace{-10pt}
\caption{\textbf{Quantitative results of generated motion and vocal.} (a) compares with different motion generation baselines, best result is shown in \textbf{bold} and the second-best result is shown in \underline{underline}. (b) compares with text-to-vocal generation baselines. $^\dag$: The MOS is calculated with 95\% confidence intervals of song samples.}
\label{tab:audiomotion}
\end{table*}

\subsection{Main Results Analysis}
\noindent\textbf{Evaluations on joint vocal and whole-body motion generations.} We compared both text-driven and audio-driven motion generation baselines in Table. \ref{tab:audiomotion} (a). 
To be noted, our setting is different from all existing methods in the following ways. First, we use rap lyrics as our textual input instead of motion textual descriptions, which contain direct action prompt words, such as \textit{walk} and \textit{jump}; Second we use text to jointly generate both audio and motion, instead of using audio to generate motion as audio-driven methods did. As is demonstrated, our model rivals with both text-to-motion and audio-to-motion methods in terms of metrics measuring body motion quality and face motion accuracy.


\begin{table*}[htbp]
\setlength{\tabcolsep}{3mm}{
\small
\centering
\begin{tabular}{l|cc|c|cc}
\toprule
Method & FID$\downarrow$ & DIV$\uparrow$ & BC$\uparrow$ & MSE$\downarrow$ & LVD$\downarrow$ \\
\midrule
GT & 0 & 14.53 & 0.499 & 0 & 0 \\
\midrule
Pretrained LLM & 48.25 & 12.65 & 0.462 & 3.28 & 12.15 \\
Single motion token & 19.15 & 12.75 & 0.477 & 2.18 & 10.12 \\
\midrule
\textbf{Ours} & \textbf{17.58} & \textbf{14.08} & \textbf{0.485} & \textbf{2.03} & \textbf{7.23}  \\
\bottomrule
\end{tabular}
\vspace{-5pt}
\caption{\textbf{Ablation study.} We compare with common designs in motion generation frameworks.}
\label{tab:ablation}
}
\end{table*}

\begin{figure}[htbp]
    \centering
    \includegraphics[width=0.8\columnwidth]{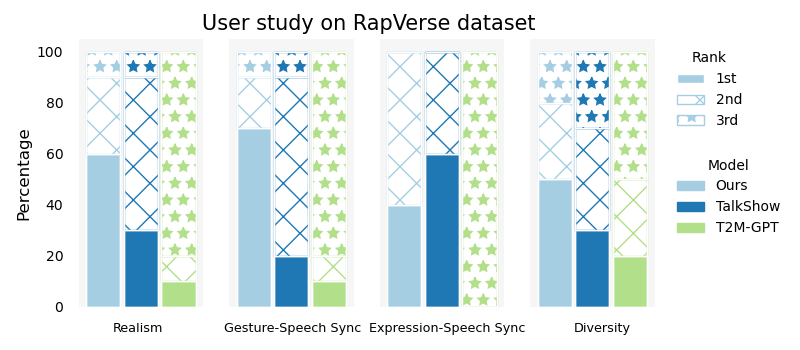}
    \vspace{-10pt}
    \caption{\textbf{User study} on generated motion.}
    \label{fig:user}
\end{figure}

Furthermore, the cornerstone of our approach lies in the simultaneous generation of vocals and motion, aiming to achieve temporal alignment between them. This objective is substantiated by our competitive results on the BC metric, which assesses the synchrony between singing vocals and corresponding motions, underscoring our success in closely synchronizing the generation of these two modalities. For the cascaded system, we integrate the text-to-vocal model DiffSinger with the audio-to-motion model Talkshow. Compared with the cascaded system, our joint-generation pipeline demonstrates superior outcomes while also reducing computational demands during both training and inference phases. In the cascaded architectures, errors tend to accumulate through each stage. Specifically, if the text-to-vocal module produces unclear vocals, it subsequently hampers the audio-to-motion model's ability to generate accurate facial expressions that align with the vocal content.

To better evaluate the generation quality, we conducted an additional user study (see Fig. \ref{fig:user}) using ten 10-second-long videos from the RapVerse dataset, sampled from the test set. We recruited 20 participants to evaluate four aspects: the holistic realism of human motion, expression-vocal synchronism, gesture-vocal synchronism, and holistic diversity, using real data as a reference. Participants were asked to sort shuffled videos generated by different methods based on these criteria. As shown in the results, our model achieved the highest scores in human motion realism, gesture-vocal synchronism, and holistic diversity when compared with T2MGPT (text-to-motion) and Talkshow (audio-to-motion).

\noindent \textbf{Evaluations on vocal generations.} We compare our method against other state-of-the-art text-to-vocal generation baselines in Table. \ref{tab:audiomotion} (b). While our unified model is trained to simultaneously generate vocals and motion, a task considerably more complex than generating vocals alone, its vocal generation component still manages to achieve results comparable to systems designed solely for vocal generations. 

\subsection{Ablation Study}
We present the outcomes of our ablation study in Table. \ref{tab:ablation}. Initially, we explored the integration of a pre-trained large language model \cite{raffel2020exploring} for multi-modality generation, akin to the approach in ~\cite{jiang2024motiongpt}. However, the efficacy of utilizing pre-trained language models significantly lags behind our tailored design, underscoring that pre-training primarily on linguistic tokens does not facilitate effective prediction across multiple modalities like vocal and motion. Additionally, we study the impact of our compositional VQ-VAEs on motion generation. In contrast, a baseline employing a single VQVAE for the joint quantization of facial, body, and hand movements was implemented. This approach led to a noticeable degradation in performance, particularly marked by a \textbf{-2.89} decrease in LVD. This decline can be attributed to the preponderance of facial movements in a singer's performance. Using a single VQ-VAE model for full-body dynamics compromises the detailed representation of facial expressions, which are crucial for realistic and coherent motion synthesis.

\section{Conclusion \label{sec:conclusion}}

In this work, we present a new framework for the simultaneous generation of 3D whole-body motions and singing vocals, directly from textual lyrics. To address this challenging task, we first collect RapVerse, a large dataset containing synchronous rap vocals, alongside lyrics and 3D whole-body motions. Utilizing RapVerse, we demonstrate that simply scaling autoregressive transformers across language, audio, and motion yields a coherent generation of singing vocals and 3D holistic human motions. We anticipate that this work will inspire novel avenues in the joint modeling of text, audio, and motion.
{
    \small
    \bibliographystyle{ieeenat_fullname}
    \bibliography{main}

@String(TOG= {ACM Trans. Graph.})

@String(ICASSP=	{ICASSP})

@String(AAAI = {AAAI})

@String(TOG   = {ACM TOG})

@article{ren2020fastspeech,
  title={Fastspeech 2: Fast and high-quality end-to-end text to speech},
  author={Ren, Yi and Hu, Chenxu and Tan, Xu and Qin, Tao and Zhao, Sheng and Zhao, Zhou and Liu, Tie-Yan},
  journal={arXiv preprint arXiv:2006.04558},
  year={2020}
}

@inproceedings{min2021meta,
  title={Meta-stylespeech: Multi-speaker adaptive text-to-speech generation},
  author={Min, Dongchan and Lee, Dong Bok and Yang, Eunho and Hwang, Sung Ju},
  booktitle={International Conference on Machine Learning},
  pages={7748--7759},
  year={2021},
  organization={PMLR}
}

@article{donahue2020end,
  title={End-to-end adversarial text-to-speech},
  author={Donahue, Jeff and Dieleman, Sander and Bi{\'n}kowski, Miko{\l}aj and Elsen, Erich and Simonyan, Karen},
  journal={arXiv preprint arXiv:2006.03575},
  year={2020}
}

@article{ren2019fastspeech,
  title={Fastspeech: Fast, robust and controllable text to speech},
  author={Ren, Yi and Ruan, Yangjun and Tan, Xu and Qin, Tao and Zhao, Sheng and Zhao, Zhou and Liu, Tie-Yan},
  journal={Advances in neural information processing systems},
  volume={32},
  year={2019}
}

@article{oord2016wavenet,
  title={Wavenet: A generative model for raw audio},
  author={Oord, Aaron van den and Dieleman, Sander and Zen, Heiga and Simonyan, Karen and Vinyals, Oriol and Graves, Alex and Kalchbrenner, Nal and Senior, Andrew and Kavukcuoglu, Koray},
  journal={arXiv preprint arXiv:1609.03499},
  year={2016}
}

@inproceedings{ren2020deepsinger,
  title={Deepsinger: Singing voice synthesis with data mined from the web},
  author={Ren, Yi and Tan, Xu and Qin, Tao and Luan, Jian and Zhao, Zhou and Liu, Tie-Yan},
  booktitle={Proceedings of the 26th ACM SIGKDD International Conference on Knowledge Discovery \& Data Mining},
  pages={1979--1989},
  year={2020}
}

@article{zhang2023t2m,
  title={T2m-gpt: Generating human motion from textual descriptions with discrete representations},
  author={Zhang, Jianrong and Zhang, Yangsong and Cun, Xiaodong and Huang, Shaoli and Zhang, Yong and Zhao, Hongwei and Lu, Hongtao and Shen, Xi},
  journal={arXiv preprint arXiv:2301.06052},
  year={2023}
}

@article{lu2023humantomato,
  title={Humantomato: Text-aligned whole-body motion generation},
  author={Lu, Shunlin and Chen, Ling-Hao and Zeng, Ailing and Lin, Jing and Zhang, Ruimao and Zhang, Lei and Shum, Heung-Yeung},
  journal={arXiv preprint arXiv:2310.12978},
  year={2023}
}

@article{spleeter2020,
  doi = {10.21105/joss.02154},
  url = {https://doi.org/10.21105/joss.02154},
  year = {2020},
  publisher = {The Open Journal},
  volume = {5},
  number = {50},
  pages = {2154},
  author = {Romain Hennequin and Anis Khlif and Felix Voituret and Manuel Moussallam},
  title = {Spleeter: a fast and efficient music source separation tool with pre-trained models},
  journal = {Journal of Open Source Software},
  note = {Deezer Research}
}

@inproceedings{heigold2016end,
  title={End-to-end text-dependent speaker verification},
  author={Heigold, Georg and Moreno, Ignacio and Bengio, Samy and Shazeer, Noam},
  booktitle={2016 IEEE International Conference on Acoustics, Speech and Signal Processing (ICASSP)},
  pages={5115--5119},
  year={2016},
  organization={IEEE}
}

@article{raffel2020exploring,
  title={Exploring the limits of transfer learning with a unified text-to-text transformer},
  author={Raffel, Colin and Shazeer, Noam and Roberts, Adam and Lee, Katherine and Narang, Sharan and Matena, Michael and Zhou, Yanqi and Li, Wei and Liu, Peter J},
  journal={Journal of machine learning research},
  volume={21},
  number={140},
  pages={1--67},
  year={2020}
}

@article{dhariwal2020jukebox,
  title={Jukebox: A generative model for music},
  author={Dhariwal, Prafulla and Jun, Heewoo and Payne, Christine and Kim, Jong Wook and Radford, Alec and Sutskever, Ilya},
  journal={arXiv preprint arXiv:2005.00341},
  year={2020}
}

@inproceedings{guo2022generating,
  title={Generating diverse and natural 3d human motions from text},
  author={Guo, Chuan and Zou, Shihao and Zuo, Xinxin and Wang, Sen and Ji, Wei and Li, Xingyu and Cheng, Li},
  booktitle={Proceedings of the IEEE/CVF Conference on Computer Vision and Pattern Recognition},
  pages={5152--5161},
  year={2022}
}

@inproceedings{li2021ai,
  title={Ai choreographer: Music conditioned 3d dance generation with aist++},
  author={Li, Ruilong and Yang, Shan and Ross, David A and Kanazawa, Angjoo},
  booktitle={Proceedings of the IEEE/CVF International Conference on Computer Vision},
  pages={13401--13412},
  year={2021}
}

@article{zhang2024motiondiffuse,
  title={Motiondiffuse: Text-driven human motion generation with diffusion model},
  author={Zhang, Mingyuan and Cai, Zhongang and Pan, Liang and Hong, Fangzhou and Guo, Xinying and Yang, Lei and Liu, Ziwei},
  journal={IEEE Transactions on Pattern Analysis and Machine Intelligence},
  year={2024},
  publisher={IEEE}
}

@article{jiang2024motiongpt,
  title={Motiongpt: Human motion as a foreign language},
  author={Jiang, Biao and Chen, Xin and Liu, Wen and Yu, Jingyi and Yu, Gang and Chen, Tao},
  journal={Advances in Neural Information Processing Systems},
  volume={36},
  year={2024}
}

@inproceedings{ahuja2019language2pose,
  title={Language2pose: Natural language grounded pose forecasting},
  author={Ahuja, Chaitanya and Morency, Louis-Philippe},
  booktitle={2019 International Conference on 3D Vision (3DV)},
  pages={719--728},
  year={2019},
  organization={IEEE}
}

@article{lu2023co,
  title={Co-Speech Gesture Synthesis using Discrete Gesture Token Learning},
  author={Lu, Shuhong and Yoon, Youngwoo and Feng, Andrew},
  journal={arXiv preprint arXiv:2303.12822},
  year={2023}
}

@inproceedings{ginosar2019learning,
  title={Learning individual styles of conversational gesture},
  author={Ginosar, Shiry and Bar, Amir and Kohavi, Gefen and Chan, Caroline and Owens, Andrew and Malik, Jitendra},
  booktitle={Proceedings of the IEEE/CVF Conference on Computer Vision and Pattern Recognition},
  pages={3497--3506},
  year={2019}
}

@inproceedings{petrovich2022temos,
  title={TEMOS: Generating diverse human motions from textual descriptions},
  author={Petrovich, Mathis and Black, Michael J and Varol, G{\"u}l},
  booktitle={European Conference on Computer Vision},
  pages={480--497},
  year={2022},
  organization={Springer}
}

@inproceedings{ghosh2021synthesis,
  title={Synthesis of compositional animations from textual descriptions},
  author={Ghosh, Anindita and Cheema, Noshaba and Oguz, Cennet and Theobalt, Christian and Slusallek, Philipp},
  booktitle={Proceedings of the IEEE/CVF international conference on computer vision},
  pages={1396--1406},
  year={2021}
}

@inproceedings{duan2013nus,
  title={The NUS sung and spoken lyrics corpus: A quantitative comparison of singing and speech},
  author={Duan, Zhiyan and Fang, Haotian and Li, Bo and Sim, Khe Chai and Wang, Ye},
  booktitle={2013 Asia-Pacific Signal and Information Processing Association Annual Summit and Conference},
  pages={1--9},
  year={2013},
  organization={IEEE}
}

@article{sharma2021nhss,
  title={NHSS: A speech and singing parallel database},
  author={Sharma, Bidisha and Gao, Xiaoxue and Vijayan, Karthika and Tian, Xiaohai and Li, Haizhou},
  journal={Speech Communication},
  volume={133},
  pages={9--22},
  year={2021},
  publisher={Elsevier}
}

@article{tamaru2020jvs,
  title={Jvs-music: Japanese multispeaker singing-voice corpus},
  author={Tamaru, Hiroki and Takamichi, Shinnosuke and Tanji, Naoko and Saruwatari, Hiroshi},
  journal={arXiv preprint arXiv:2001.07044},
  year={2020}
}

@article{ogawa2021tohoku,
  title={Tohoku Kiritan singing database: A singing database for statistical parametric singing synthesis using Japanese pop songs},
  author={Ogawa, Itsuki and Morise, Masanori},
  journal={Acoustical Science and Technology},
  volume={42},
  number={3},
  pages={140--145},
  year={2021},
  publisher={Acoustical Society of Japan}
}

@inproceedings{liu2022diffsinger,
  title={Diffsinger: Singing voice synthesis via shallow diffusion mechanism},
  author={Liu, Jinglin and Li, Chengxi and Ren, Yi and Chen, Feiyang and Zhao, Zhou},
  booktitle={Proceedings of the AAAI conference on artificial intelligence},
  volume={36},
  number={10},
  pages={11020--11028},
  year={2022}
}

@inproceedings{xing2023codetalker,
  title={Codetalker: Speech-driven 3d facial animation with discrete motion prior},
  author={Xing, Jinbo and Xia, Menghan and Zhang, Yuechen and Cun, Xiaodong and Wang, Jue and Wong, Tien-Tsin},
  booktitle={Proceedings of the IEEE/CVF Conference on Computer Vision and Pattern Recognition},
  pages={12780--12790},
  year={2023}
}

@inproceedings{huang2021multi,
  title={Multi-singer: Fast multi-singer singing voice vocoder with a large-scale corpus},
  author={Huang, Rongjie and Chen, Feiyang and Ren, Yi and Liu, Jinglin and Cui, Chenye and Zhao, Zhou},
  booktitle={Proceedings of the 29th ACM International Conference on Multimedia},
  pages={3945--3954},
  year={2021}
}

@article{wang2022opencpop,
  title={Opencpop: A high-quality open source chinese popular song corpus for singing voice synthesis},
  author={Wang, Yu and Wang, Xinsheng and Zhu, Pengcheng and Wu, Jie and Li, Hanzhao and Xue, Heyang and Zhang, Yongmao and Xie, Lei and Bi, Mengxiao},
  journal={arXiv preprint arXiv:2201.07429},
  year={2022}
}

@article{zhang2022m4singer,
  title={M4singer: A multi-style, multi-singer and musical score provided mandarin singing corpus},
  author={Zhang, Lichao and Li, Ruiqi and Wang, Shoutong and Deng, Liqun and Liu, Jinglin and Ren, Yi and He, Jinzheng and Huang, Rongjie and Zhu, Jieming and Chen, Xiao and others},
  journal={Advances in Neural Information Processing Systems},
  volume={35},
  pages={6914--6926},
  year={2022}
}

@article{plappert2016kit,
  title={The KIT motion-language dataset},
  author={Plappert, Matthias and Mandery, Christian and Asfour, Tamim},
  journal={Big data},
  volume={4},
  number={4},
  pages={236--252},
  year={2016},
  publisher={Mary Ann Liebert, Inc. 140 Huguenot Street, 3rd Floor New Rochelle, NY 10801 USA}
}

@inproceedings{mahmood2019amass,
  title={AMASS: Archive of motion capture as surface shapes},
  author={Mahmood, Naureen and Ghorbani, Nima and Troje, Nikolaus F and Pons-Moll, Gerard and Black, Michael J},
  booktitle={Proceedings of the IEEE/CVF international conference on computer vision},
  pages={5442--5451},
  year={2019}
}

@article{lin2024motion,
  title={Motion-x: A large-scale 3d expressive whole-body human motion dataset},
  author={Lin, Jing and Zeng, Ailing and Lu, Shunlin and Cai, Yuanhao and Zhang, Ruimao and Wang, Haoqian and Zhang, Lei},
  journal={Advances in Neural Information Processing Systems},
  volume={36},
  year={2024}
}

@inproceedings{yi2023generating,
  title={Generating holistic 3d human motion from speech},
  author={Yi, Hongwei and Liang, Hualin and Liu, Yifei and Cao, Qiong and Wen, Yandong and Bolkart, Timo and Tao, Dacheng and Black, Michael J},
  booktitle={Proceedings of the IEEE/CVF Conference on Computer Vision and Pattern Recognition},
  pages={469--480},
  year={2023}
}

@inproceedings{cao2017realtime,
  title={Realtime multi-person 2d pose estimation using part affinity fields},
  author={Cao, Zhe and Simon, Tomas and Wei, Shih-En and Sheikh, Yaser},
  booktitle={Proceedings of the IEEE conference on computer vision and pattern recognition},
  pages={7291--7299},
  year={2017}
}

@inproceedings{habibie2021learning,
  title={Learning speech-driven 3d conversational gestures from video},
  author={Habibie, Ikhsanul and Xu, Weipeng and Mehta, Dushyant and Liu, Lingjie and Seidel, Hans-Peter and Pons-Moll, Gerard and Elgharib, Mohamed and Theobalt, Christian},
  booktitle={Proceedings of the 21st ACM International Conference on Intelligent Virtual Agents},
  pages={101--108},
  year={2021}
}

@article{liu2023emage,
  title={EMAGE: Towards Unified Holistic Co-Speech Gesture Generation via Masked Audio Gesture Modeling},
  author={Liu, Haiyang and Zhu, Zihao and Becherini, Giorgio and Peng, Yichen and Su, Mingyang and Zhou, You and Iwamoto, Naoya and Zheng, Bo and Black, Michael J},
  journal={arXiv preprint arXiv:2401.00374},
  year={2023}
}

@inproceedings{cudeiro2019capture,
  title={Capture, learning, and synthesis of 3D speaking styles},
  author={Cudeiro, Daniel and Bolkart, Timo and Laidlaw, Cassidy and Ranjan, Anurag and Black, Michael J},
  booktitle={Proceedings of the IEEE/CVF Conference on Computer Vision and Pattern Recognition},
  pages={10101--10111},
  year={2019}
}

@article{wuu2022multiface,
  title={Multiface: A dataset for neural face rendering},
  author={Wuu, Cheng-hsin and Zheng, Ningyuan and Ardisson, Scott and Bali, Rohan and Belko, Danielle and Brockmeyer, Eric and Evans, Lucas and Godisart, Timothy and Ha, Hyowon and Huang, Xuhua and others},
  journal={arXiv preprint arXiv:2207.11243},
  year={2022}
}

@inproceedings{takeuchi2017creating,
  title={Creating a gesture-speech dataset for speech-based automatic gesture generation},
  author={Takeuchi, Kenta and Kubota, Souichirou and Suzuki, Keisuke and Hasegawa, Dai and Sakuta, Hiroshi},
  booktitle={HCI International 2017--Posters' Extended Abstracts: 19th International Conference, HCI International 2017, Vancouver, BC, Canada, July 9--14, 2017, Proceedings, Part I 19},
  pages={198--202},
  year={2017},
  organization={Springer}
}

@inproceedings{ferstl2018investigating,
  title={Investigating the use of recurrent motion modelling for speech gesture generation},
  author={Ferstl, Ylva and McDonnell, Rachel},
  booktitle={Proceedings of the 18th International Conference on Intelligent Virtual Agents},
  pages={93--98},
  year={2018}
}

@article{taylor2017deep,
  title={A deep learning approach for generalized speech animation},
  author={Taylor, Sarah and Kim, Taehwan and Yue, Yisong and Mahler, Moshe and Krahe, James and Rodriguez, Anastasio Garcia and Hodgins, Jessica and Matthews, Iain},
  journal={ACM Transactions On Graphics (TOG)},
  volume={36},
  number={4},
  pages={1--11},
  year={2017},
  publisher={ACM New York, NY, USA}
}

@article{shen2023difftalk,
  title={Difftalk: Crafting diffusion models for generalized talking head synthesis},
  author={Shen, Shuai and Zhao, Wenliang and Meng, Zibin and Li, Wanhua and Zhu, Zheng and Zhou, Jie and Lu, Jiwen},
  journal={arXiv preprint arXiv:2301.03786},
  year={2023}
}

@inproceedings{zhang2023sadtalker,
  title={SadTalker: Learning Realistic 3D Motion Coefficients for Stylized Audio-Driven Single Image Talking Face Animation},
  author={Zhang, Wenxuan and Cun, Xiaodong and Wang, Xuan and Zhang, Yong and Shen, Xi and Guo, Yu and Shan, Ying and Wang, Fei},
  booktitle={Proceedings of the IEEE/CVF Conference on Computer Vision and Pattern Recognition},
  pages={8652--8661},
  year={2023}
}

@article{fanelli20103,
  title={A 3-d audio-visual corpus of affective communication},
  author={Fanelli, Gabriele and Gall, Juergen and Romsdorfer, Harald and Weise, Thibaut and Van Gool, Luc},
  journal={IEEE Transactions on Multimedia},
  volume={12},
  number={6},
  pages={591--598},
  year={2010},
  publisher={IEEE}
}

@article{hennequin2020spleeter,
  title={Spleeter: a fast and efficient music source separation tool with pre-trained models},
  author={Hennequin, Romain and Khlif, Anis and Voituret, Felix and Moussallam, Manuel},
  journal={Journal of Open Source Software},
  volume={5},
  number={50},
  pages={2154},
  year={2020}
}

@article{redmon2018yolov3,
  title={Yolov3: An incremental improvement},
  author={Redmon, Joseph and Farhadi, Ali},
  journal={arXiv preprint arXiv:1804.02767},
  year={2018}
}

@article{schneider2023mo,
  title={Mo$\backslash$\^{} usai: Text-to-music generation with long-context latent diffusion},
  author={Schneider, Flavio and Kamal, Ojasv and Jin, Zhijing and Sch{\"o}lkopf, Bernhard},
  journal={arXiv preprint arXiv:2301.11757},
  year={2023}
}

@misc{polyak2021speech,
      title={Speech Resynthesis from Discrete Disentangled Self-Supervised Representations}, 
      author={Adam Polyak and Yossi Adi and Jade Copet and Eugene Kharitonov and Kushal Lakhotia and Wei-Ning Hsu and Abdelrahman Mohamed and Emmanuel Dupoux},
      year={2021},
      eprint={2104.00355},
      archivePrefix={arXiv},
      primaryClass={cs.SD}
}

@article{agostinelli2023musiclm,
  title={Musiclm: Generating music from text},
  author={Agostinelli, Andrea and Denk, Timo I and Borsos, Zal{\'a}n and Engel, Jesse and Verzetti, Mauro and Caillon, Antoine and Huang, Qingqing and Jansen, Aren and Roberts, Adam and Tagliasacchi, Marco and others},
  journal={arXiv preprint arXiv:2301.11325},
  year={2023}
}

@article{huang2023noise2music,
  title={Noise2music: Text-conditioned music generation with diffusion models},
  author={Huang, Qingqing and Park, Daniel S and Wang, Tao and Denk, Timo I and Ly, Andy and Chen, Nanxin and Zhang, Zhengdong and Zhang, Zhishuai and Yu, Jiahui and Frank, Christian and others},
  journal={arXiv preprint arXiv:2302.03917},
  year={2023}
}

@article{van2017neural,
  title={Neural discrete representation learning},
  author={Van Den Oord, Aaron and Vinyals, Oriol and others},
  journal={Advances in neural information processing systems},
  volume={30},
  year={2017}
}

@inproceedings{pavlakos2019expressive,
  title={Expressive body capture: 3d hands, face, and body from a single image},
  author={Pavlakos, Georgios and Choutas, Vasileios and Ghorbani, Nima and Bolkart, Timo and Osman, Ahmed AA and Tzionas, Dimitrios and Black, Michael J},
  booktitle={Proceedings of the IEEE/CVF conference on computer vision and pattern recognition},
  pages={10975--10985},
  year={2019}
}

@inproceedings{valin2019lpcnet,
  title={LPCNet: Improving neural speech synthesis through linear prediction},
  author={Valin, Jean-Marc and Skoglund, Jan},
  booktitle={ICASSP 2019-2019 IEEE International Conference on Acoustics, Speech and Signal Processing (ICASSP)},
  pages={5891--5895},
  year={2019},
  organization={IEEE}
}

@inproceedings{teed2020raft,
  title={Raft: Recurrent all-pairs field transforms for optical flow},
  author={Teed, Zachary and Deng, Jia},
  booktitle={Computer Vision--ECCV 2020: 16th European Conference, Glasgow, UK, August 23--28, 2020, Proceedings, Part II 16},
  pages={402--419},
  year={2020},
  organization={Springer}
}

@inproceedings{yang2021multi,
  title={Multi-band melgan: Faster waveform generation for high-quality text-to-speech},
  author={Yang, Geng and Yang, Shan and Liu, Kai and Fang, Peng and Chen, Wei and Xie, Lei},
  booktitle={2021 IEEE Spoken Language Technology Workshop (SLT)},
  pages={492--498},
  year={2021},
  organization={IEEE}
}

@inproceedings{radford2023robust,
  title={Robust speech recognition via large-scale weak supervision},
  author={Radford, Alec and Kim, Jong Wook and Xu, Tao and Brockman, Greg and McLeavey, Christine and Sutskever, Ilya},
  booktitle={International Conference on Machine Learning},
  pages={28492--28518},
  year={2023},
  organization={PMLR}
}

@inproceedings{hsu2021hubert,
  title={HuBERT: How much can a bad teacher benefit ASR pre-training?},
  author={Hsu, Wei-Ning and Tsai, Yao-Hung Hubert and Bolte, Benjamin and Salakhutdinov, Ruslan and Mohamed, Abdelrahman},
  booktitle={ICASSP 2021-2021 IEEE International Conference on Acoustics, Speech and Signal Processing (ICASSP)},
  pages={6533--6537},
  year={2021},
  organization={IEEE}
}

@inproceedings{kasi2002yet,
  title={Yet another algorithm for pitch tracking},
  author={Kasi, Kavita and Zahorian, Stephen A},
  booktitle={2002 ieee international conference on acoustics, speech, and signal processing},
  volume={1},
  pages={I--361},
  year={2002},
  organization={IEEE}
}

@article{zhang2023stylesinger,
  title={StyleSinger: Style Transfer for Out-Of-Domain Singing Voice Synthesis},
  author={Zhang, Yu and Huang, Rongjie and Li, Ruiqi and He, JinZheng and Xia, Yan and Chen, Feiyang and Duan, Xinyu and Huai, Baoxing and Zhao, Zhou},
  journal={arXiv preprint arXiv:2312.10741},
  year={2023}
}

@article{zhang2022wesinger,
  title={Wesinger: Data-augmented singing voice synthesis with auxiliary losses},
  author={Zhang, Zewang and Zheng, Yibin and Li, Xinhui and Lu, Li},
  journal={arXiv preprint arXiv:2203.10750},
  year={2022}
}

@article{he2023rmssinger,
  title={RMSSinger: Realistic-Music-Score based Singing Voice Synthesis},
  author={He, Jinzheng and Liu, Jinglin and Ye, Zhenhui and Huang, Rongjie and Cui, Chenye and Liu, Huadai and Zhao, Zhou},
  journal={arXiv preprint arXiv:2305.10686},
  year={2023}
}

@inproceedings{guo2022tm2t,
  title={Tm2t: Stochastic and tokenized modeling for the reciprocal generation of 3d human motions and texts},
  author={Guo, Chuan and Zuo, Xinxin and Wang, Sen and Cheng, Li},
  booktitle={European Conference on Computer Vision},
  pages={580--597},
  year={2022},
  organization={Springer}
}

@inproceedings{chen2023executing,
  title={Executing your Commands via Motion Diffusion in Latent Space},
  author={Chen, Xin and Jiang, Biao and Liu, Wen and Huang, Zilong and Fu, Bin and Chen, Tao and Yu, Gang},
  booktitle={Proceedings of the IEEE/CVF Conference on Computer Vision and Pattern Recognition},
  pages={18000--18010},
  year={2023}
}

@inproceedings{kim2023flame,
  title={Flame: Free-form language-based motion synthesis \& editing},
  author={Kim, Jihoon and Kim, Jiseob and Choi, Sungjoon},
  booktitle={Proceedings of the AAAI Conference on Artificial Intelligence},
  volume={37},
  number={7},
  pages={8255--8263},
  year={2023}
}

@article{tevet2022human,
  title={Human motion diffusion model},
  author={Tevet, Guy and Raab, Sigal and Gordon, Brian and Shafir, Yonatan and Cohen-Or, Daniel and Bermano, Amit H},
  journal={arXiv preprint arXiv:2209.14916},
  year={2022}
}

@article{kim2023muse,
  title={MuSE-SVS: Multi-Singer Emotional Singing Voice Synthesizer that Controls Emotional Intensity},
  author={Kim, Sungjae and Kim, Yewon and Jun, Jewoo and Kim, Injung},
  journal={IEEE/ACM Transactions on Audio, Speech, and Language Processing},
  year={2023},
  publisher={IEEE}
}

@book{lakoff1980metaphors,
  author    = {George Lakoff and Mark Johnson},
  title     = {Metaphors We Live By},
  year      = {1980},
  publisher = {University of Chicago Press},
  edition   = {2003},
  isbn      = {9780226468013},
  language  = {English},
  pages     = {276},
  note      = {First published January 1, 1980},
  url       = {https://www.goodreads.com/book/show/3869.Metaphors_We_Live_By},
}

@article{bian2024adding,
  title={Adding Multi-modal Controls to Whole-body Human Motion Generation},
  author={Bian, Yuxuan and Zeng, Ailing and Ju, Xuan and Liu, Xian and Zhang, Zhaoyang and Liu, Wei and Xu, Qiang},
  journal={arXiv preprint arXiv:2407.21136},
  year={2024}
}

@inproceedings{chen2024diffsheg,
  title={Diffsheg: A diffusion-based approach for real-time speech-driven holistic 3d expression and gesture generation},
  author={Chen, Junming and Liu, Yunfei and Wang, Jianan and Zeng, Ailing and Li, Yu and Chen, Qifeng},
  booktitle={Proceedings of the IEEE/CVF Conference on Computer Vision and Pattern Recognition},
  pages={7352--7361},
  year={2024}
}

@article{alexanderson2023listen,
  title={Listen, denoise, action! audio-driven motion synthesis with diffusion models},
  author={Alexanderson, Simon and Nagy, Rajmund and Beskow, Jonas and Henter, Gustav Eje},
  journal={ACM Transactions on Graphics (TOG)},
  volume={42},
  number={4},
  pages={1--20},
  year={2023},
  publisher={ACM New York, NY, USA}
}
}

\end{document}